\documentclass[letterpaper, 10 pt, conference]{ieeeconf}  

\IEEEoverridecommandlockouts                              
\usepackage[pdftex]{graphicx}

\let\labelindent\relax

\usepackage{array}
\usepackage{amsfonts}
\usepackage{times}
\usepackage{enumitem}
\usepackage[ruled,linesnumbered]{algorithm2e}
\usepackage{algorithmic}
\usepackage[normalem]{ulem}
\usepackage{amsmath}
\usepackage{amsmath,amssymb,amsbsy}
\usepackage{breqn}
\usepackage{xcolor} 
\usepackage{comment}
\usepackage{enumerate}
\usepackage{epstopdf}
\usepackage{textcomp}
\usepackage{hyperref}
\usepackage{subfig}

\usepackage[utf8]{inputenc}
\usepackage[T1]{fontenc}
\usepackage{nomencl}
\makenomenclature
\newcommand{\vect}[1]{\boldsymbol{#1}}

\definecolor{darkcyan}{rgb}{0.0, 0.55, 0.55}
\newcommand{\spr}[1] {{\color{red} #1}}
\newcommand{\azy}[1] {{\color{blue} #1}}

\begin{document}

\title{
Configuration Tracking Control of a Multi-Segment Soft Robotic Arm Using a Cosserat Rod Model}
\author{Azadeh~Doroudchi$^{*1}$, Zhi Qiao$^{*2}$, Wenlong Zhang$^{3}$, and Spring~Berman$^{2}$ %
\thanks{This work was supported by 
Office of Naval Research (ONR) Award N00014-17-1-2117 and  National Science Foundation (NSF) Award 
CMMI-1800940. }
\thanks{${}^{*}_{}$These authors contributed equally to the paper.}
\thanks{$^{1}$Azadeh Doroudchi is with School of Electrical, Computer and Energy Engineering, Arizona State University, Tempe, AZ, 85287, USA
        {\tt\footnotesize adoroudc@asu.edu}} 
\thanks{$^{2}$Zhi Qiao and Spring Berman are with the School for Engineering of Matter, Transport and Energy, Arizona State University, Tempe, AZ, 85287, USA
        {\tt\footnotesize zqiao7@asu.edu, spring.berman@asu.edu}}        
\thanks{$^{3}$Wenlong Zhang is with the School of Manufacturing Systems and Networks, Fulton Schools of Engineering, Arizona State University, Mesa, AZ, 85212, USA
        {\tt\footnotesize wenlong.zhang@asu.edu}}
}

\maketitle
\begin{abstract}
Controlling soft continuum robotic arms is challenging due to their hyper-redundancy and dexterity. In this paper we demonstrate, for the first time, closed-loop 
control of the configuration space variables of a soft 
robotic arm, composed of independently controllable 
segments, 
using a Cosserat rod model of the robot and the distributed sensing and actuation capabilities of the segments.
%
%
Our controller solves the inverse dynamic problem by simulating the Cosserat rod model in MATLAB using a computationally efficient numerical solution scheme, and it applies the computed control output to the actual robot in real time. 
%
%
The position and orientation of the tip of each segment are measured in real time, while the remaining unknown variables that are needed to solve the inverse dynamics are estimated simultaneously in the simulation. We implement the controller on a multi-segment silicone robotic arm with pneumatic actuation, using a motion capture system to measure the segments' positions and orientations. The controller is used to reshape the 
arm into configurations that are achieved through different combinations of bending and extension deformations in 3D space. The resulting tracking performance indicates the effectiveness of the controller 
and the accuracy of the simulated Cosserat rod model that is used to estimate the unmeasured variables.
\end{abstract}


\section{Introduction}

\begin{figure}[t]
\centering
\includegraphics[width=0.45\textwidth]{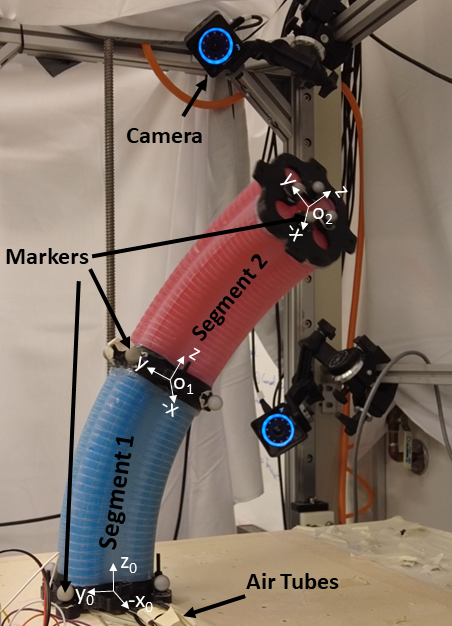}
    \caption{The multi-segment silicone
    robotic arm used in this work. The silicone segments were fabricated using a method similar to the one in~\cite{nguyen2017design}.}
    \label{fig:multi-segment-silicone-arm}
        \vspace{-0.4in}
\end{figure}

Soft continuum robots have many potential uses in manipulation and locomotion tasks that require high dexterity and compliance, and have often been inspired by soft biological structures with these properties, such as octopus arms and elephant trunks~\cite{kapadia2011task, calisti2011octopus}. The control of soft robotic arms is a challenging problem due to the robots' inherent passive compliance, infinite degrees of freedom, and nonlinear material characteristics~\cite{george2018control,jiang2021hierarchical}. Substantial progress has been made on this problem in recent years \cite{chen2022review,schegg2022review,mengaldo2022concise}, and the development of soft segments with independent actuation~\cite{della2018dynamic} and soft grippers with integrated sensors~\cite{homberg2019robust} are expanding the autonomous capabilities of soft robots in unstructured environments.

Both model-based and model-free approaches have been used to design controllers for soft robotic arms~\cite{george2018control}. Of the model-based approaches, dynamic controllers are generally more accurate than kinematic controllers; however, the high computational complexity and sensing requirements of existing dynamic controllers have limited their usage~\cite{webster_2010, katzschmann2019dynamic}. The first closed-loop dynamic controller for a soft continuum robot was developed in~\cite{della2020model} and experimentally validated on a 2D soft robotic arm, composed of multiple segments with pneumatic actuation, for trajectory tracking and surface following tasks. The inverse dynamics solution was obtained by adding a mixed feedforward–feedback term to the closed-loop controller based on the Lagrangian formulation of the robot 
dynamics. Another model-based 
dynamic controller was designed in~\cite{chang2021controlling} to control a simulated octopus arm with realistic muscular actuation for reaching and grasping tasks. The octopus arm was modeled as a Cosserat rod, which can describe large deformations due to bending, torsion, shear, and extension~\cite{antman2005problems}. In~\cite{boyer2011macrocontinuous}, continuum robots inspired by elongated body animals (e.g., snakes, worms, and caterpillars) were modeled as Kirchhoff rods, a special case of the Cosserat rod model, and used to solve the inverse dynamics problem to simulate terrestrial locomotion gaits. However, it is unclear how to implement this approach in practice on an underactuated continuum robot~\cite{till2019real}.

In our previous work~\cite{doroudchi2021configuration}, we proposed a new model-based inverse dynamic control approach for reshaping a soft robotic arm, modeled as a Cosserat rod, consisting of independently-controllable segments with local sensing and actuation. The controller drives the robot to target 3D configurations through bending, torsion, shear, and extension deformations. We validated the controller in numerical simulations of robotic arms composed of either hydrogel or silicone. In this paper, we adapt our controller from~\cite{doroudchi2021configuration} to a soft robotic arm composed of pneumatically-actuated silicone segments, shown in Fig.~\ref{fig:multi-segment-silicone-arm}, and experimentally validate the controller on this platform. To enable the implementation of the controller, we define mappings from the robot's joint space to its actuator space (desired air pressures) and from its task space to its configuration space (curvatures, extension). The inverse dynamics of the robot are solved using real-time measurements of the positions and orientations of its segments, which are obtained by a motion capture system, and estimates of unmeasured variables from the solution to the forward dynamics of the Cosserat rod model.

In summary, the main contributions of this paper are the following: 
\begin{description}[font=$\bullet$]
    \item Design of a Cosserat rod model-based, inverse dynamic control approach for 3D configuration tracking by a soft robotic arm composed of independently-controllable segments with pneumatic actuation
    \item Estimation of unmeasured robot variables in the inverse dynamics solution using real-time simulation of the Cosserat rod model
    \item Experimental validation of the control approach on a multi-segment silicone robotic arm that is capable of bending and extending in 3D space
\end{description}

The remainder of the paper is organized as follows. Section~\ref{sec:Silicone Segment} 
describes the design, fabrication, and physical properties of the pneumatically-actuated silicone segments. In Section~\ref{sec:dynamics}, the forward and inverse dynamics of the robotic arm are derived based on a Cosserat rod model of the robot. Then, the configuration tracking controller and numerical solution of the forward and inverse dynamics are outlined in Section~\ref{sec:control}. The controller's tracking performance in both simulations and experiments is discussed in Section~\ref{sec:simulation}. Finally, Section~\ref{sec:conclusion} concludes and suggests directions for future work.

\section{Silicone Segment Fabrication and Properties}
\label{sec:Silicone Segment}

\begin{figure}[t]
\centering
\vspace{3pt}
\includegraphics[width=0.45\textwidth]{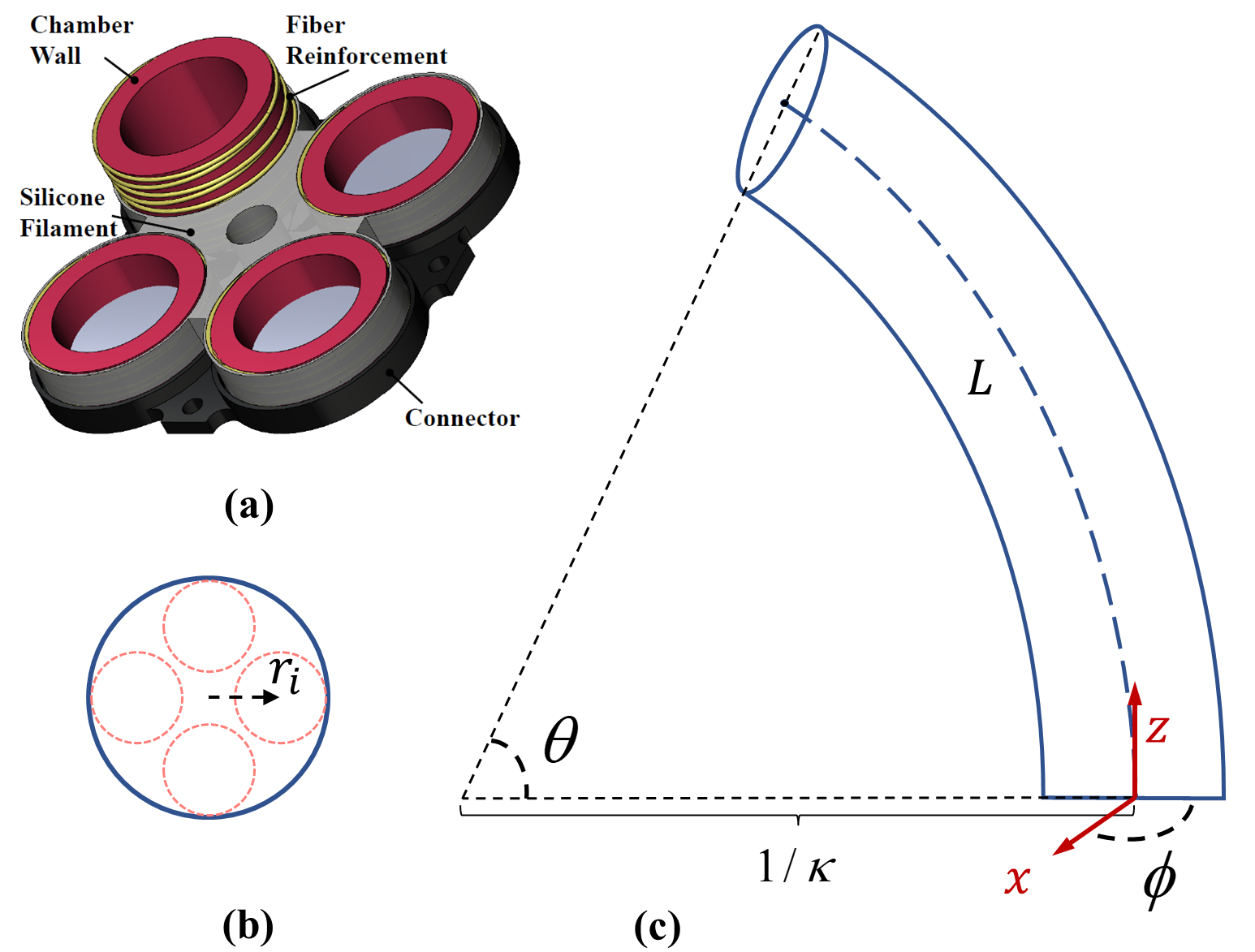} 
    \caption{(a) Design of a silicone segment. 
    (b) Cross-section of a segment. $\vect{r}_i$ is the vector from the backbone to the center of the $i$-th FRA chamber in the local frame of the cross-section. The cross-section has area $A$, and the $i$-th FRA chamber has cap area $A_i$.
    (c) Kinematic representation of a constant-curvature segment with respect to a local frame fixed at the base of the segment.
    } 
    \label{fig:PCC}
    \vspace{-0.2in}
\end{figure}



The multi-segment silicone arm was fabricated using a process similar to the one described in~\cite{nguyen2017design}. Figure~\ref{fig:PCC}a illustrates the design of a silicone segment, Fig.~\ref{fig:PCC}b shows its cross-section, and Fig.~\ref{fig:PCC}c depicts its kinematic representation as a constant-curvature segment. As shown in Fig.~\ref{fig:PCC}a, each segment contains four fiber-reinforced actuators (FRAs). These four FRAs are symmetrically arranged with respect to the central axis of the segment and embedded in the segment using silicone filament. 
The helical fiber reinforcements are wound clockwise and counterclockwise in such a way that the FRAs elongate along the axial direction instead of expanding in the radial direction.
As shown in Fig.~\ref{fig:multi-segment-silicone-arm}, the two segments are dyed either blue or red and are connected by black 3D-printed parts, which are fastened with bolts and nuts.
Compressed air is applied to each segment through four air tubes, each of which is connected to an FRA and a digital pressure regulator. The pressure regulator measures the air pressure $p_m$ in the FRA and drives it to a desired pressure set-point $p_d$.



\begin{table}[t]
\centering
\caption{Silicone segment properties.}
\begin{tabular}{l l l l}
\hline
Param. & Description & Value & Units \\
\hline
$N$ & Number of segments & 2 & -- \\
$r_i$ & Distance from backbone to each FRA  & 0.3 & cm\\
$A_i$ & FRA cap area & 3.1 & cm$^2$ \\
$r_0$ & Undeformed radius of the segment & 5.3 & cm\\
$L_0$ & Undeformed length of the segment & 18.5& cm\\
$M$ & Mass of the segment & 0.825 & kg\\
$\rho$ & Density of the segment & 792.8 & kg$/$m$^3$\\
$G$ & Shear modulus of the segment & 0.1 & MPa \\
$E$ & Young's modulus of the segment & 0.28 & MPa \\
\hline
\end{tabular}
\label{table:param-silicone}
\vspace{-0.1in} 
\end{table}

The physical properties of each segment are listed in Table~\ref{table:param-silicone}. In order to estimate the Young's modulus of a segment, assuming that its value remains constant during actuation, an Instron 5944 universal testing machine was used to elongate the unactuated segment under three loading speeds (1 mm/s, 3 mm/s and 5 mm/s) and measure its corresponding change in length and the applied force. Three trials were conducted for each loading speed, and the Young's modulus was estimated as the average value over all nine trials. 


\section{Dynamics and Kinematics of the Robot Arm 
based on the Cosserat Rod Model}
\label{sec:dynamics}






In this section, we develop a Cosserat rod model of the nonlinear dynamics of the soft robotic arm. This model is accurate under the assumptions of a sufficiently large length-to-radius ratio, material incompressibility, and linear elasticity. 

\subsection{Cosserat rod model}
\label{subsec:dynamics}
The robot arm is modeled as an elastic Cosserat rod of length $L$ and density $\rho$. Each of its cross-sections has area $A$ and second mass moment of inertia tensor $\vect{J}$. The cross-section at arc length $s$ in the global coordinate frame $G$ has position ${}^{G}_{}\vect{p}(t,s) \in \mathbb{R}^{3}$ and orientation matrix ${}^{G}_{}\vect{R}(t,s) \in$ \text{\emph{SO}(3)} at time $t$ (see Fig. 1 in~\cite{doroudchi2021configuration} for a schematic of a Cosserat rod in Cartesian coordinates). From here on, a variable without the annotation $G$ is defined with respect to a local coordinate frame that is fixed to the cross-section with which it is associated.

The configuration space variables of the rod, given that its neutral axis is in the $z$ direction, are defined as the curvature vector, $\vect{u}(t,s)=[u_x,u_y,u_z]^T$, and the rate of change of position,  $\vect{v}(t,s)=[v_x,v_y,v_z]^T$. Here, the components $u_x$ and $u_y$ produce bending about the $x$ and $y$ axes; $u_z$ produces torsion about the $z$-axis; $v_x$ and $v_y$ cause shear effects that change the size of the cross-section; and $v_z$ produces extension along the $z$-axis. 
The vectors $\vect{q}(t,s)$ and $\vect{w}(t,s)$ denote the translational and angular velocities, respectively, of the cross-section at arc length $s$. The force and moment that the material at $\vect{p}(t,s+ds)$ exerts on the material at $\vect{p}(t,s-ds)$, for infinitesimal $ds$, are called the {\it internal} force and moment, ${}^{G}_{}\vect{n}(t,s)\in\mathbb{R}^{3}$ and ${}^{G}_{}\vect{m}(t,s)\in~\mathbb{R}^{3}$. Any force and moment that are applied to the backbone are called an {\it external} force and moment, ${}^{G}_{}\vect{f}(t,s)\in\mathbb{R}^{3}$ and ${}^{G}_{}\vect{l}(t,s)\in\mathbb{R}^{3}$.

The deformation of each cross-section of the rod is governed by a set of partial differential equations, differentiated with respect to $s$ and $t$. The spatial derivatives of the state variables are calculated at each cross-section. Defining {$\widehat{(\cdot)}$} as the cross product matrix of a vector, the internal force and moment evolve according to the equations:
\begin{equation}
\label{Eq_PDE_ID}
\begin{split}
{}^{G}\vect{n}_s&= {}^{G}\vect{R}\rho A(\widehat{\vect{w}}\vect{q}+\vect{q}_t)-{}^{G}\vect{f}, \\
{}^{G}\vect{m}_s&= {}^{G}\vect{R}\rho(\widehat{\vect{w}}\vect{J}\vect{w}+\vect{J}\vect{w}_t)-{}^{G}\widehat{\vect{p}}_s{}^{G}\vect{n}-{}^{G}\vect{l},
\end{split}
\end{equation}
\noindent and the kinematic variables evolve according to: 
\begin{equation}
\label{Eq_PDE_IK}
\begin{split}
{}^{G}\vect{p}_s&={}^{G}\vect{R}\vect{v} ~, {}^{G}\vect{p}_t=\vect{R}\vect{q},\\
{}^{G}\vect{R}_s&={}^{G}\vect{R}\vect{\widehat{u}} ~, {}^{G}\vect{R}_t=\vect{R}\vect{\widehat{w}},\\
\vect{q}_s&=\vect{v}_t-\widehat{\vect{u}}\vect{q}+\widehat{\vect{w}}\vect{v}, \\
\vect{w}_s&=\vect{u}_t-\widehat{\vect{u}}\vect{w}.
\end{split}
\end{equation}
The time derivatives are computed using the Backward Differentiation Formula (BDF)~\cite{stoer2013introduction,schaller2019robotic}.


\subsection{Joint space to actuator space mapping}
\label{subsec:J2A}
The sources of the external forces and moments are the control inputs applied by the robot's pneumatic actuators, which produce force ${}^{G}_{}\vect{f}_p$ and moment ${}^{G}_{}\vect{l}_p$, and the gravitational force per unit length  in the global frame, ${}^{G}_{}\vect{f}_e$:
\begin{equation}
\begin{split}
\label{Eq_force-moment-silicone}
{}^{G}\vect{f}&={}^{G}\vect{f}_{p}+ {}^{G}\vect{f}_{e},\\ 
{}^{G}\vect{l}&={}^{G}\vect{l}_{p}, 
\end{split}
\end{equation}
where
\begin{equation}
\label{Eq_force-gravity-silicone}
{}^{G}\vect{f}_{e}=\rho A {}^{G}\vect{g}, ~~~~ {}^{G}\vect{g}=[0~0~-9.81]^T\,m/s^2.
\end{equation}
The forces and moments applied by the pneumatic actuators to the backbone are given by: 
\begin{equation}
\begin{split}
\label{Eq_force-moment-p}
{}^{G}\vect{f}_{p}&=\sum_{i=1}^{4} p_i A_i[{}^{G}\vect{R}_s\vect{e}_3]-\rho A {}^{G}\vect{g},\\ 
{}^{G}\vect{l}_{p}&=\sum_{i=1}^{4} p_i A_i\dfrac{\partial{}}{\partial{s}}[({}^{G}\vect{p}+{}^{G}\vect{R}\vect{r}_i)\times {}^{G}\vect{R}\vect{e}_3],
\end{split}
\end{equation}
where $p_i$ is the chamber air pressure of the $i$-th FRA and, as depicted in Fig.~\ref{fig:PCC}b, $A_i$ is the corresponding chamber cap area, $\vect{r}_i$ is the vector from the center of the backbone to the center of the $i$-th FRA in the local frame attached to a segment cross-section, and $\vect{e}_3$ is the unit vector along the $z$-axis. The gravitational effect in~\eqref{Eq_force-gravity-silicone} is subtracted from the actuation force to cancel out its effect on the backbone in~\eqref{Eq_force-moment-silicone}.

Each silicone segment has three DOFs: bending about the $x$-axis, bending about the $y$-axis, and elongation along the $z$-axis. Hence, the equivalent actuation for each segment, $\vect{P}=[P_x, P_y, P_z]$, can be computed using the mapping below:
\begin{equation}
\label{Eq:mapping-pd-1}
\begin{split}
P_x&={}^{G}\vect{l}_p(y) \cdot {}^{G}\vect{l}^T_p(y) /A_i,\\ 
P_y&={}^{G}\vect{l}_p(x) \cdot {}^{G}\vect{l}^T_p(x) /A_i,\\ 
P_z&={}^{G}\vect{f}_p(z) \cdot {}^{G}\vect{f}^T_p(z) /A_i, \\
\end{split}
\end{equation}
where $A_i$ is equal for all the chambers. The equivalent actuation is related to the real actuator pressures, $p_1, p_2, p_3, p_4$, as follows:
\begin{equation}
\label{Eq:mapping-pd-2}
\begin{split}
{P_x}&=-p_1+p_2-p_3+p_4,\\
{P_y}&=p_1+p_2-p_3-p_4,\\
{P_z}&=p_1+p_2+p_3+p_4.
\end{split}
\end{equation}
Since our silicone arm is similar in design to the Honeycomb Pneumatic Networks (HPN) Arm \cite{jiang2021hierarchical}, in which each segment has two bending DOFs and one elongation DOF, the relation $p_1+p_4=p_2+p_3$ must hold. Therefore, the real actuator pressures are computed as:
\begin{equation}
\label{Eq:mapping-pd-3}
\begin{split}
p_1&=(-P_x+P_y+P_z)/4,\\
p_2&=(P_x+P_y+P_z)/4,\\
p_3&=(-P_x-P_y+P_z)/4,\\
p_4&=(P_x-P_y+P_z)/4.
\end{split}
\end{equation}

\subsection{Forward dynamics solution}
\label{subsec:FD}
The solution to the forward dynamics of the rod is obtained by substituting \eqref{Eq:mapping-pd-3} into \eqref{Eq_force-moment-p}, \eqref{Eq_force-gravity-silicone} and \eqref{Eq_force-moment-p} into \eqref{Eq_force-moment-silicone}, and \eqref{Eq_force-moment-silicone} into \eqref{Eq_PDE_ID}, and then spatially integrating \eqref{Eq_PDE_ID} over the length of the rod to calculate the configuration space variables as follows:
\begin{equation}
\label{Eq_FD}
\begin{split}  
\vect{v}&=(\vect{K}_{se}+c_0\vect{B}_{se})^{-1} [{}^{G}\vect{R}^T~{}^{G}\vect{n}+\vect{K}_{se}\vect{v}^*-\vect{B}_{se}\vect{v}_h],\\
\vect{u}&=(\vect{K}_{bt}+c_0\vect{B}_{bt})^{-1} [{}^{G}\vect{R}^T~{}^{G}\vect{m}+\vect{K}_{bt}\vect{u}^*-\vect{B}_{bt}\vect{u}_h].
\end{split}
\end{equation}
Here, $\vect{v}^*$ and $\vect{u}^*$ are the vectors $\vect{v}$ and $\vect{u}$ at the undeformed reference shape, and the history elements $\vect{v}_h$ and $\vect{u}_h$ are calculated from the values of $\vect{v}$ and $\vect{u}$ at the previous two time steps~\cite{till2019real}. The vectors $\vect{v}_h$ and $\vect{u}_h$ are obtained from experimental data on the position and orientation of the robot in its task space, as discussed in Section \ref{sec:control}. The matrices $\vect{K}_{se}$, $\vect{K}_{bt}$, $\vect{B}_{se}$, and $\vect{B}_{bt}$ in \eqref{Eq_FD} are defined as:
\begin{eqnarray}
\label{Eq_stiffness}
\vect{K}_{se}&=\begin{bmatrix}  \alpha_c G & 0  & 0 \\ 0 & \alpha_c G & 0 \\ 0 & 0 & E
\end{bmatrix} A , \quad \vect{K}_{bt}= \begin{bmatrix}  E & 0  & 0 \\ 0 & E & 0 \\ 0 & 0 & G
\end{bmatrix} \vect{J}, \nonumber \\
\vect{B}_{se}& \hspace{-4.2cm} = \tau \vect{K}_{se}, \quad \vect{B}_{bt}= \tau \vect{K}_{bt}, 
\end{eqnarray}
where $G$ and $E$ are the shear modulus and Young's modulus, respectively, of the silicone segment (given in Table \ref{table:param-silicone}) and $\alpha_c = 4/3$
for circular cross-sections. The damping matrices $\vect{B}_{se}$ and $\vect{B}_{bt}$ are calculated from vibration tests~\cite{linn2013geometrically}, in which $\tau$ is twice the period of vibrations exhibited by the robot arm's tip. Due to the high stiffness of the segments, we set $\tau\approx0$.

\subsection{Task space to configuration space mapping}
\label{subsec:T2C}
Since the configuration space variables ($u_x$, $u_y$, $v_z$) cannot be directly measured from the robot's task space, the solution to the inverse dynamics problem requires a mapping from the robot's bending and extension deformations in the task space to the corresponding values of $u_x$, $u_y$, and $v_z$. Since this mapping is robot-independent~\cite{webster_2010}, we use the Piecewise Constant Curvature (PCC) configuration space variables, ($\kappa(t,s),\phi(t,s),L(t,s)$), to complete the mapping. These variables are defined for a constant-curvature segment that approximates the silicone segment, illustrated in Fig.~\ref{fig:PCC}c: $\kappa$ is the curvature of the segment's backbone, $L$ is its length, and $\phi$ is its angle of rotation about the $z$-axis with respect to the $x$-axis (positive counterclockwise). The position ($x,y,z$) of the tip of the segment with respect to the local frame attached to its base is related to the corresponding PCC configuration space variables as follows, defined as in \cite{jiang2021hierarchical}:
\begin{equation}
\label{Eq_PCC_kapa}
\kappa= \begin{cases}
2x/(x^2+z^2), \quad \quad \phi=0 \text{ rad}\\ 
2y/(y^2+z^2), \quad \quad \phi=\pi/2 \text{ rad}\\ 
0, \quad \quad \quad \quad \quad \quad \quad x=0, ~y=0
\end{cases}
\end{equation}

\begin{equation}
\label{Eq_PCC_l}
L= \begin{cases}
(1/\kappa) \arctan \left(x/z\right), \quad \quad \phi=0 \text{ rad}\\ 
(1/\kappa) \arctan \left(y/z\right), \quad  \quad \phi=\pi/2 \text{ rad}\\ 
z, \quad  \quad \quad \quad \qquad  \quad \quad \quad ~ x=0, ~y=0
\end{cases}
\end{equation}
The segment's backbone lies in the $x$--$z$ plane at $\phi=0$ rad and in the $y$--$z$ plane at $\phi=\pi/2$ rad.
The curvature variable $\kappa$ in each of these planes is equal to the component of the curvature vector $\vect{u}$ in the same plane for bending deformations, and the extension variable $v_z$ is given by the extension ratio of the segment:
\begin{equation}
\label{Eq_PCC}
\begin{split}
u_y&= \kappa \quad \text{for} \quad \phi=0 \text{ rad},\\
u_x&= -\kappa \quad \text{for} \quad \phi=\pi/2 \text{ rad},\\ 
v_z&=1+\dfrac{L-L_0}{L_0},
\end{split}
\end{equation}
where $L_0$ is the undeformed length of the segment. The inverse dynamics solution can be obtained once the control input is computed, which is discussed in the following section.

\section{Inverse Dynamic Control of the Robot Arm}
\label{sec:control}

\begin{figure*}[pt]
\centering
\includegraphics[width=0.87\textwidth]{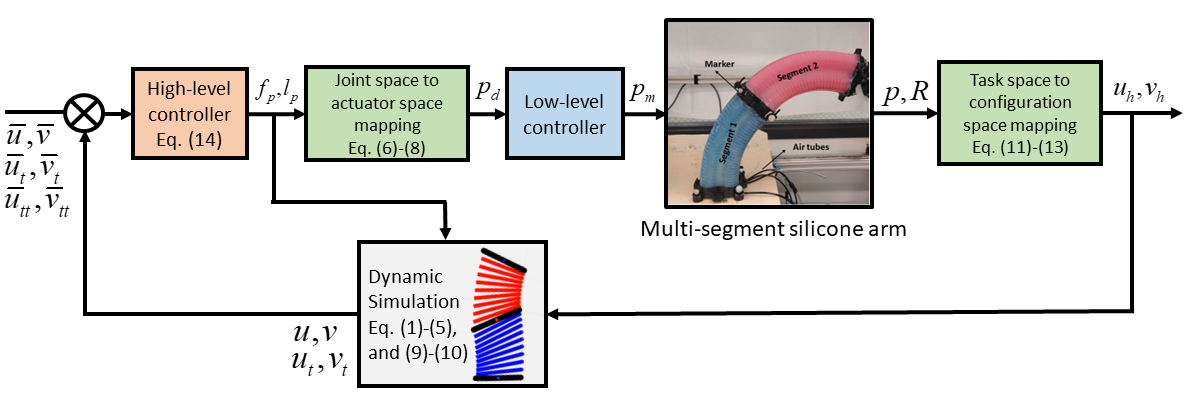} 
    \caption{Schematic block diagram of the configuration tracking controller.} 
    \label{fig:hybrid-control}
    \vspace{-0.1in} 
\end{figure*}

In this section, we design a control approach that drives the soft multi-segment robotic arm to track a time-varying reference configuration denoted by ($\vect{\bar{u}}(t), \vect{\bar{v}}(t)$), as well as its first and second time derivatives, that achieves desired bending and extension deformations in 3D space.
We specify a {\it high-level controller} that is similar to the dynamic controller in our prior work~\cite{doroudchi2021configuration},
with proportional-derivative gain matrices that
are defined in terms of the physical and material properties of distinct cross-sections of the robot arm. The outputs of the high-level controller are defined as:
\begin{equation}
\label{Eq_control_law-silicone}
\begin{split}  
{}^{G}\vect{f}_{\it p} = &{}^{G}\vect{R} \big[\vect{K}_{p_1}(\vect{\bar{v}}-\vect{v})+\vect{K}_{v_1}(\vect{\bar{v}}_t-\vect{v}_t)+\vect{K}_{m_1}\vect{\bar{v}}_{tt}\big] \\ & -\rho A~ {}^{G}\vect{g}, \\
{}^{G}\vect{l}_{\it p}= &{}^{G}\vect{R} \big[\vect{K}_{p_2}(\vect{\bar{u}}-\vect{u})+\vect{K}_{v_2}(\vect{\bar{u}}_t-\vect{u}_t)+\vect{K}_{m_2}\vect{\bar{u}}_{tt}\big],  
\end{split}
\end{equation}
where $\vect{K}_{m_1}$ and $\vect{K}_{m_2}$ are $3\times 3$ diagonal matrices whose diagonal entries are proportional to $\rho A$ and $\rho \vect{J}$, respectively, and $\vect{K}_{p_1}, \vect{K}_{p_2}, \vect{K}_{v_1}, \vect{K}_{v_2}$ are the diagonal gain matrices 
\begin{equation}
\label{Eq_matrices_KK}
\begin{split}
\vect{K}_{p_1}&=p_1\vect{K}_{se},\quad \vect{K}_{v_1}=v_1\vect{B}_{se},\\
\vect{K}_{p_2}&=p_2\vect{K}_{bt},\quad \vect{K}_{v_2}=v_2\vect{B}_{bt},
\end{split}
\end{equation}
\noindent where we manually tune the 
coefficients to ${v_1}=1, ~{v_2}=1$ for both segments, ${p_1}=100, ~{p_2}=50$ for segment 1, and ${p_1}=37.5,~ {p_2}=18.75$ for segment 2. The closed-loop system is proven to be globally asymptotically stable in the Appendix. 


Figure~\ref{fig:hybrid-control} illustrates a schematic block diagram of our configuration tracking controller. 
The reference configuration ($\vect{\bar{u}}(t), \vect{\bar{v}}(t)$) and its first and second time derivatives are  sent to the control loop at each time step. 
The high-level controller computes the control outputs, which are the desired forces and moments ($\vect{f}_p, \vect{l}_p$) applied by the actuators, according to~\eqref{Eq_control_law-silicone} based on the difference between these reference values and the current configuration ($\vect{u}(t), \vect{v}(t)$) and their time derivatives, computed numerically for the simulated robot. These control outputs are mapped to the desired air pressure $p_d$ in each FRA using~\eqref{Eq:mapping-pd-1}-\eqref{Eq:mapping-pd-3}. Then, the {\it low-level controller} drives the measured air pressure $p_m$ in each FRA of the real robot to the corresponding desired set-point $p_d$ via the robot's pneumatic actuators. After the robot responds to the actuation, the position and orientation of the tip of each segment ($\vect{p}, \vect{R}$) are measured by an Optitrack motion capture system with Motive software. Using the mapping~\eqref{Eq_PCC_kapa}-\eqref{Eq_PCC}, the measured task space variables ($\vect{p}, \vect{R}$) are used to compute the configuration space variables ($\vect{u}, \vect{v}$), which are stored as the history elements ($\vect{u}_h, \vect{v}_h$) for the next two time steps. Lastly, to close the loop, the values of ($\vect{u}, \vect{v}$) and their first time derivatives at the next time step are approximated by the simulation of the Cosserat model forward dynamics based on~\eqref{Eq_PDE_ID}-\eqref{Eq_force-moment-p},~\eqref{Eq_FD}-\eqref{Eq_stiffness}, using the high-level controller output \eqref{Eq_control_law-silicone}. 

\begin{algorithm}[t]
\caption{Configuration tracking controller 
}
\label{alg:dynamic-silicone}
\begin{algorithmic}[1]
	    \STATE Given~ 
	    $\vect{\bar{u}}^i_j, \vect{\bar{v}}^i_j, ~\vect{\bar{u}}^i_{t,j}, \vect{\bar{v}}^i_{t,j}, ~\vect{\bar{u}}^i_{tt,j}, \vect{\bar{v}}^i_{tt,j},$ \newline $~~~~~~~~~~ i = 1,...,T/dt, ~j = 1,...,L/ds$
    	\FOR {$i \gets 1$  to $T/dt$}
		\STATE $\vect{n}^i_0,\vect{m}^i_0 \gets$ SSM
	    ($\vect{n}^{i-1}_L=\vect{m}^{i-1}_L=\vect{0}$)
		\FOR {$j \gets 1$  to $L/ds$}
		\STATE $\vect{n}^i_{j},\vect{m}^i_{j} \gets$ RK4 using ($\vect{n}^i_{j-1},\vect{m}^i_{j-1}$) and
		($\vect{n}^i_{s,j-1},\vect{m}^i_{s,j-1}$)
		\STATE $\vect{u}^i_{j},\vect{v}^i_{j} \gets$ Forward dynamics~\eqref{Eq_FD} using $\vect{n}^i_j,\vect{m}^i_j$ and $\vect{u}^i_{h,j},\vect{v}^i_{h,j}$
		\STATE $\vect{f}^i_{p,j},\vect{l}^i_{p,j} \gets $ Control law~\eqref{Eq_control_law-silicone} using $\vect{v}^i_{j},\vect{u}^i_{j}$ and numerical approximations of their time derivatives 
		\STATE $\vect{P}^i_{j} \gets$ Eq.~\eqref{Eq:mapping-pd-1} using $\vect{f}^i_{p,j},\vect{l}^i_{p,j}$
		\STATE $\vect{n}^i_{s,j},\vect{m}^i_{s,j} \gets$ Eq. \eqref{Eq_PDE_ID} using $\vect{f}^i_{p,j},\vect{l}^i_{p,j}$ 
		\ENDFOR
		\ENDFOR \\
		\noindent Note: $(\vect{n},\vect{m},\vect{f},\vect{l})$ are defined in the global frame, and $(\vect{v},\vect{u},\vect{P})$ in the local frame.
\end{algorithmic}
\end{algorithm}

Algorithm~\ref{alg:dynamic-silicone} describes the numerical solution of the forward and inverse dynamic problems in the simulation. The numerical solution of the forward dynamics is obtained using the computationally inexpensive approach in~\cite{till2019real}. First, the time-varying reference configuration and its first and second time derivatives are defined (line 1).
An {\it outer loop} iterates over time steps $i$ (lines 2 to 11), and an {\it inner loop}
iterates over discretized spatial locations (nodes) $j$ along the backbone of the robot (lines 4 to 10). In the outer loop, the boundary conditions $\vect{n}^i_0={}^{G}\vect{n}(i,0)$ and $\vect{m}^i_0= {}^{G}\vect{m}(i,0)$ of the fixed end of the arm are guessed using the standard shooting method (SSM) (line 3), which converts a two-point boundary value problem (BVP) to an initial value problem (IVP), as described in~\cite{holsapple2003modified}, given that $\vect{n}^{i-1}_L={}^{G}\vect{n}(i-1,L) =\vect{0}$ and $\vect{m}^{i-1}_L={}^{G}\vect{m}(i-1,L)=\vect{0}$ at the free end of the arm and $\vect{n}^0_0$ and $\vect{m}^0_0$ are set to $\vect{0}$. The implicit fourth-order Runge-Kutta (RK4) method is applied to compute the internal force and moment $\vect{n}^i_{j}$, $\vect{m}^i_{j}$ from their current values and spatial derivatives at spatial node $j-1$ (line 5). Next, the values of $\vect{v}^i_j$ and $\vect{u}^i_j$ are computed according to \eqref{Eq_FD} using the numerically integrated $\vect{n}^i_j,\vect{m}^i_j$ and the history elements $\vect{v}^i_{h,j},\vect{u}^i_{h,j}$ from the past two time steps (line 6). Then, using the error between the configuration space variables and their desired values and the error between their actual and desired time derivatives, control law~\eqref{Eq_control_law-silicone} is used to calculate the force and moment ($\vect{f}^i_{p,j}$,$\vect{l}^i_{p,j}$) that the pneumatic actuators must apply to the corresponding backbone section (line 7). These forces and moments are then mapped to the equivalent actuation $\vect{P}$ (line 8), which is converted using \eqref{Eq:mapping-pd-3} into the desired actuator pressures that are sent to the actual robot.
Lastly, $\vect{n}^i_{s,j}$ and $\vect{m}^i_{s,j}$ are found for the next iteration of the inner loop (line 9). 

\begin{figure*}[t]
    \centering
    \subfloat[\centering 
    Desired, simulated, and experimental $v_z(t)$ for both segments. 
    ]{{\includegraphics[width=0.5\linewidth]{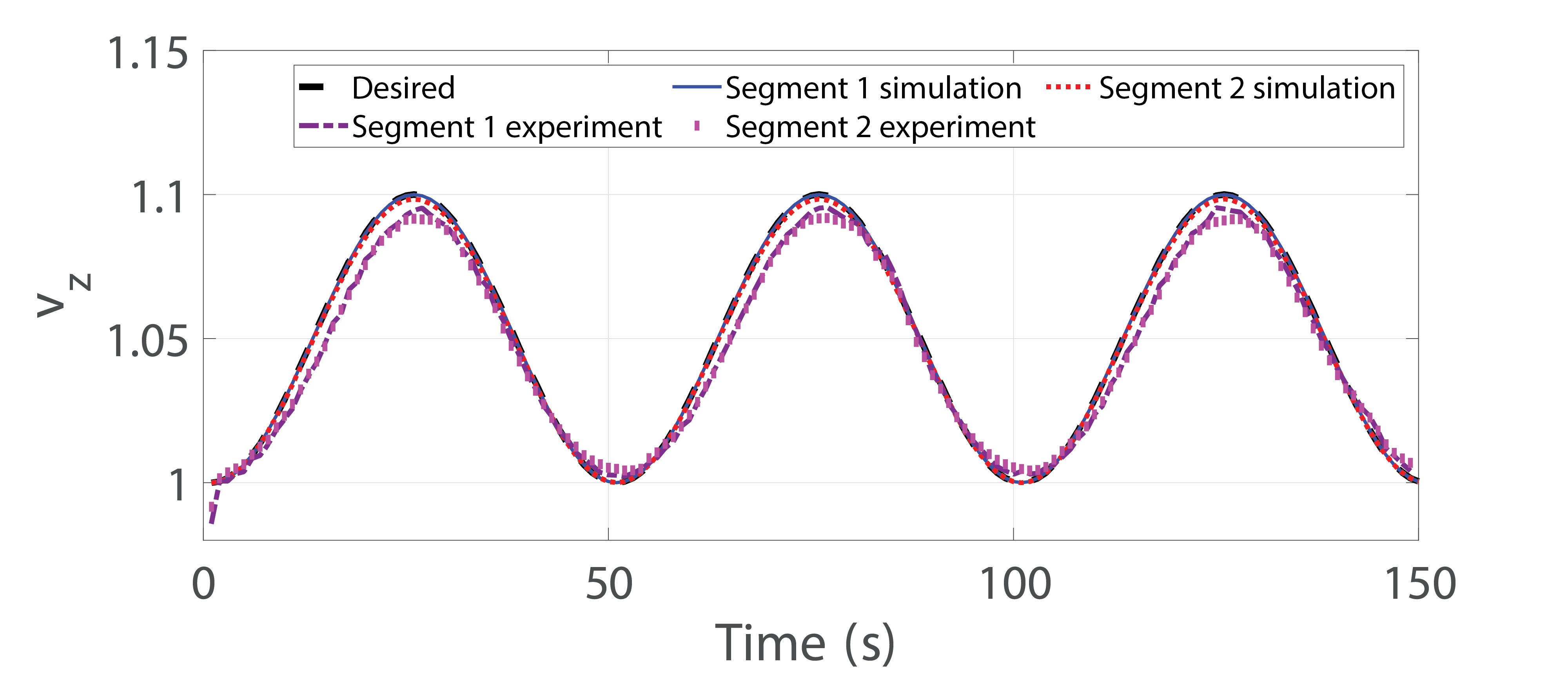} }}
    \subfloat[\centering Height $z(t)$ of segment 1's tip in simulation and experiment. 
    ]{{\includegraphics[width=0.5\linewidth]{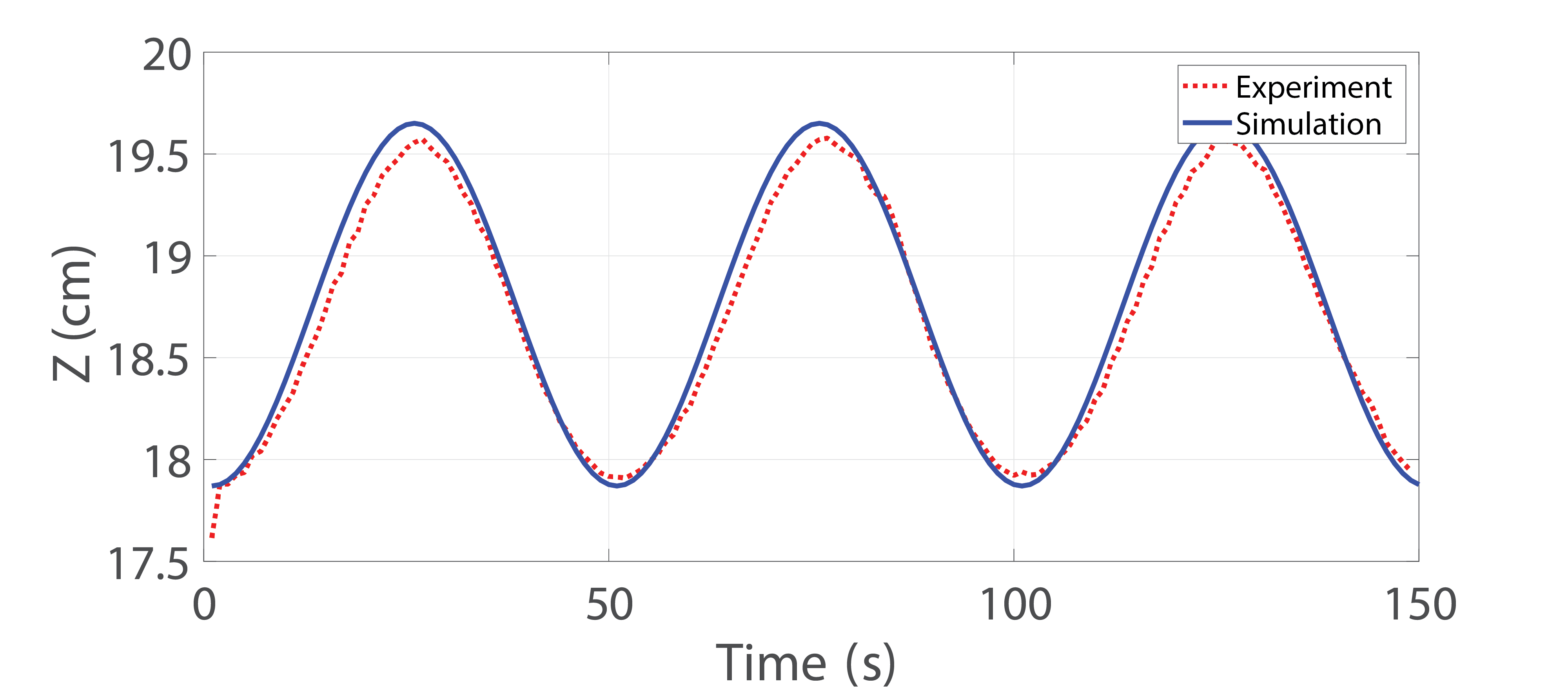} }}
    \hfill
    \subfloat[\centering Height $z(t)$ of segment 2's tip in simulation and experiment. 
    ]{{\includegraphics[width=0.5\linewidth]{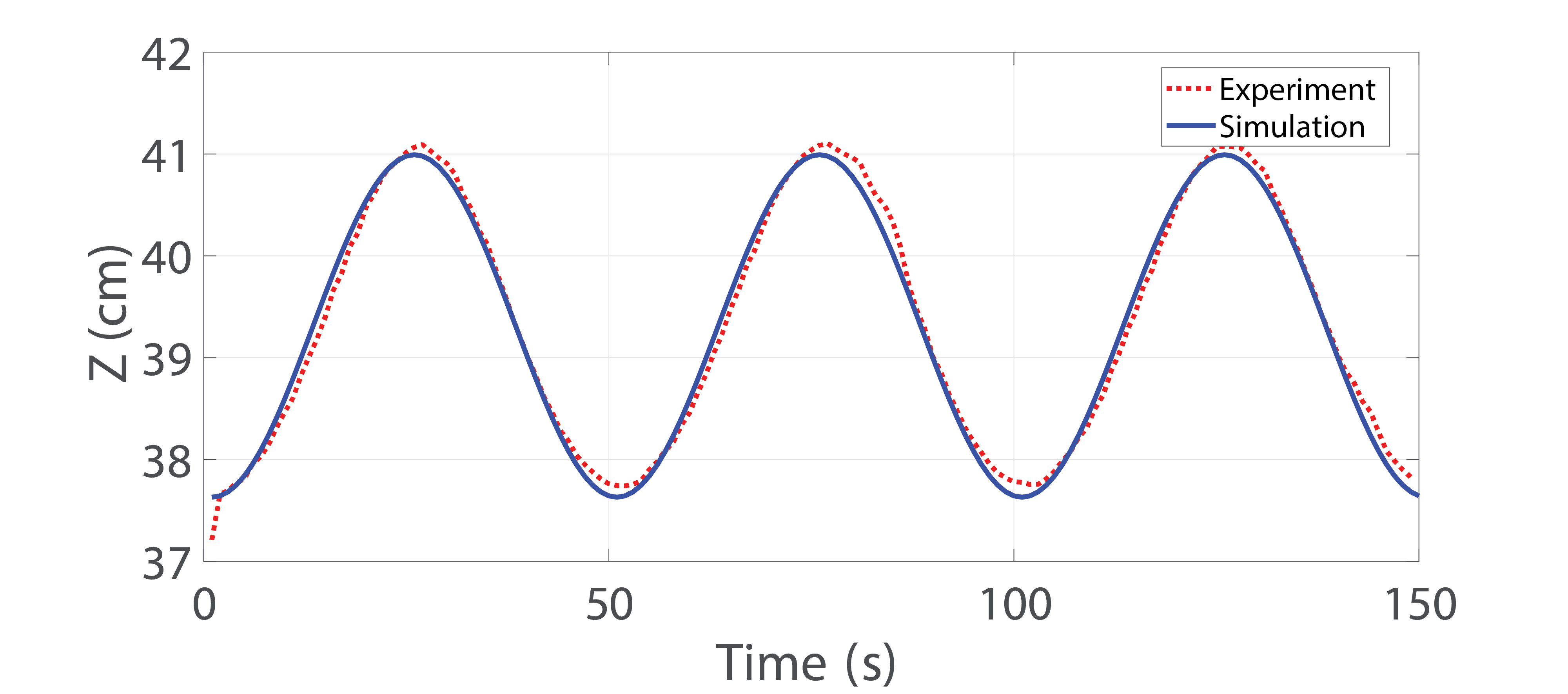} }}
    \subfloat[\centering Desired pressure $p_{d,j}^i$ and measured pressure $p_{m,j}^i$ for actuator $j$ of segment $i$ during the experiment. 
    ]{{\includegraphics[width=0.5\linewidth]{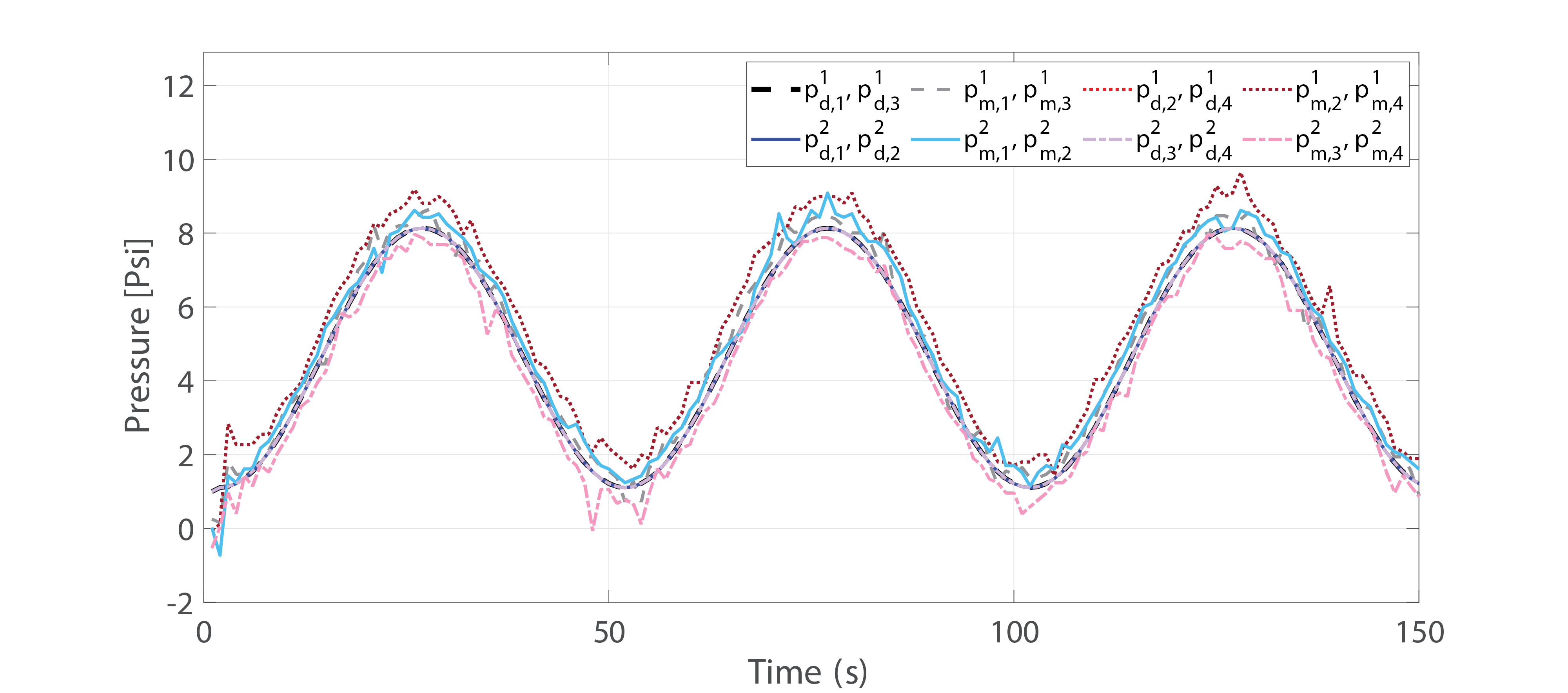} }}
    \caption{Controller performance for tracking the reference extension $\bar{v}_z(t)$ in \eqref{Eq_v_z_silicone} in an experimental trial with the soft robotic arm and the corresponding simulation of its dynamical model.
    }
    \label{fig:extension-control-1}
    \vspace{-0.2in} 
\end{figure*}

\section{Simulation and Experimental Results}
\label{sec:simulation}


In this section, we compare the performance of 
the controller at tracking 
reference configurations in both the actual robotic arm (Fig. \ref{fig:multi-segment-silicone-arm}) and the simulation of the arm, which runs simultaneously in real time. As mentioned earlier, the controller can produce any of the four main deformations; however, due to the robot's physical design restrictions, we are only able to validate the performance of the controller for bending and extension deformations. In addition, while there are no theoretical limitations imposed by our modeling and control approach on the number of segments in the arm, we have limited the number of segments to two due to space constraints on the motion capture system. The simulated arm is slender, uniform, and symmetric about the $z$-axis and has isotropic material properties. Although the simulated arm has a circular cross-section (with radius $r_0$), our control approach can also be applied to  multi-segment robots with other cross-section geometries, as long as they satisfy the assumptions required for using the Cosserat rod model.

We tested sinusoidal reference inputs with different amplitudes and frequencies that produce extension and bending deformations in the robot under open-loop actuation. During each test, the position and orientation of the tip of each segment were recorded using the motion capture cameras, and the dead-zone of the electric valve of each FRA was avoided by pre-loading the FRAs at a pressure of 1 psi. We describe results for reference inputs that kept the robot within the tracked space of the motion capture system: one that produced extension of both segments, and four that produced simultaneous bending of segment 1 about the $(+x)$- or $(-x)$-axis (i.e., toward the $(-y)$- or $(+y)$-axis) and bending of segment 2 about the $(+y)$- or $(-y)$-axis (i.e., toward the $(+x)$- or $(-x)$-axis). Video recordings of tests with these reference inputs, including  ones not discussed in this section due to space limitations, are shown in the supplementary attachment.

The reference input that produced extension was defined for both segments as:
\begin{equation}
\label{Eq_v_z_silicone}
\begin{split}  
\bar{v}_z(t)= 1 + a\sin^2\left(\omega t\right),
\end{split}
\end{equation}
where $a=0.1$ cm, $\omega=2\pi/100$ rad/s. Figures~\ref{fig:extension-control-1}a-d compare the desired, simulated, and actual extensions $v_z(t)$, 
the $z$ coordinates of the segment tips, and 
the desired and measured actuator pressures. The figures show that the robot closely tracks the reference input $\bar{v}_z(t)$, and that the simulation accurately predicts the robot's deformation over time. 

\begin{figure*}[t]
    \centering
    \subfloat[\centering Desired, simulated, and experimental $u_x(t)$ for segment 1.
    ]{{\includegraphics[width=0.5\linewidth]{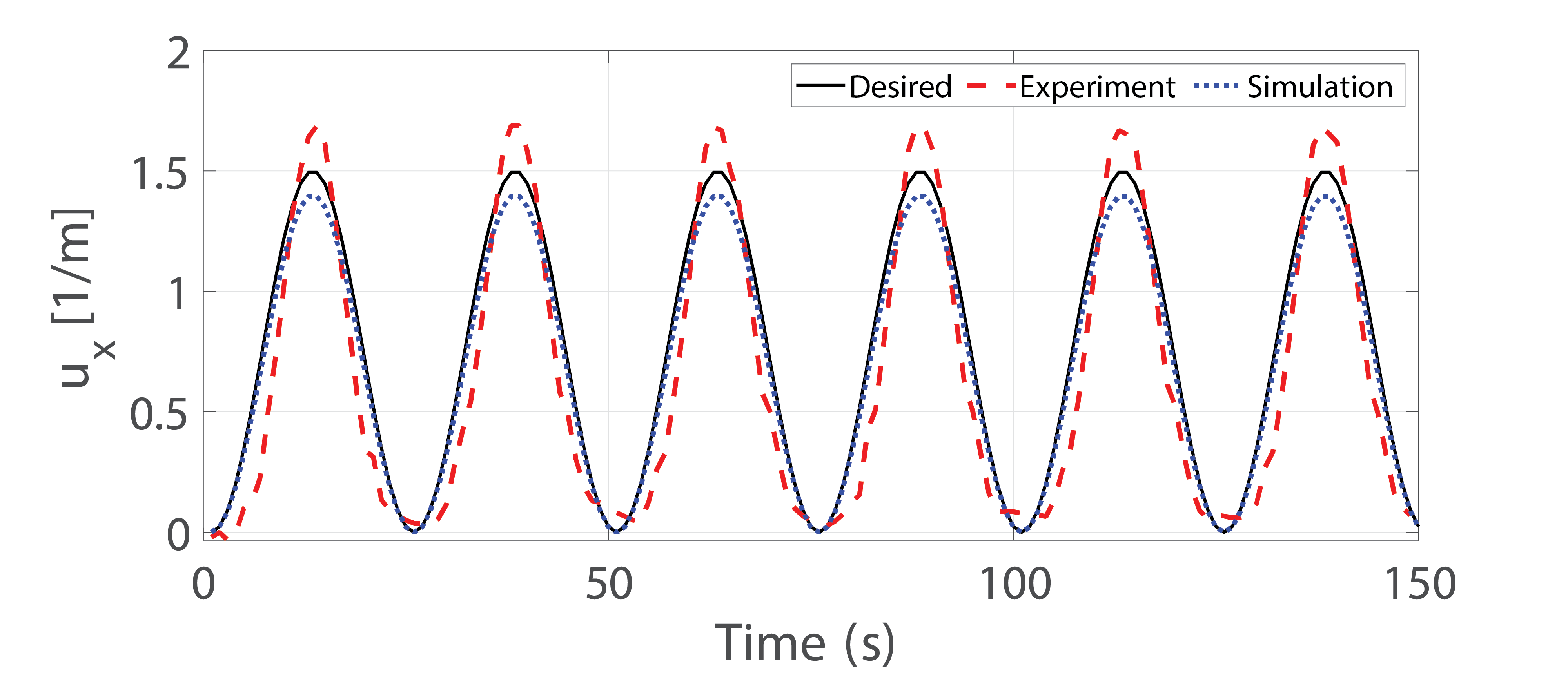} }}
    \subfloat[\centering Desired, simulated, and experimental $u_y(t)$ for segment 2.
    ]{{\includegraphics[width=0.5\linewidth]{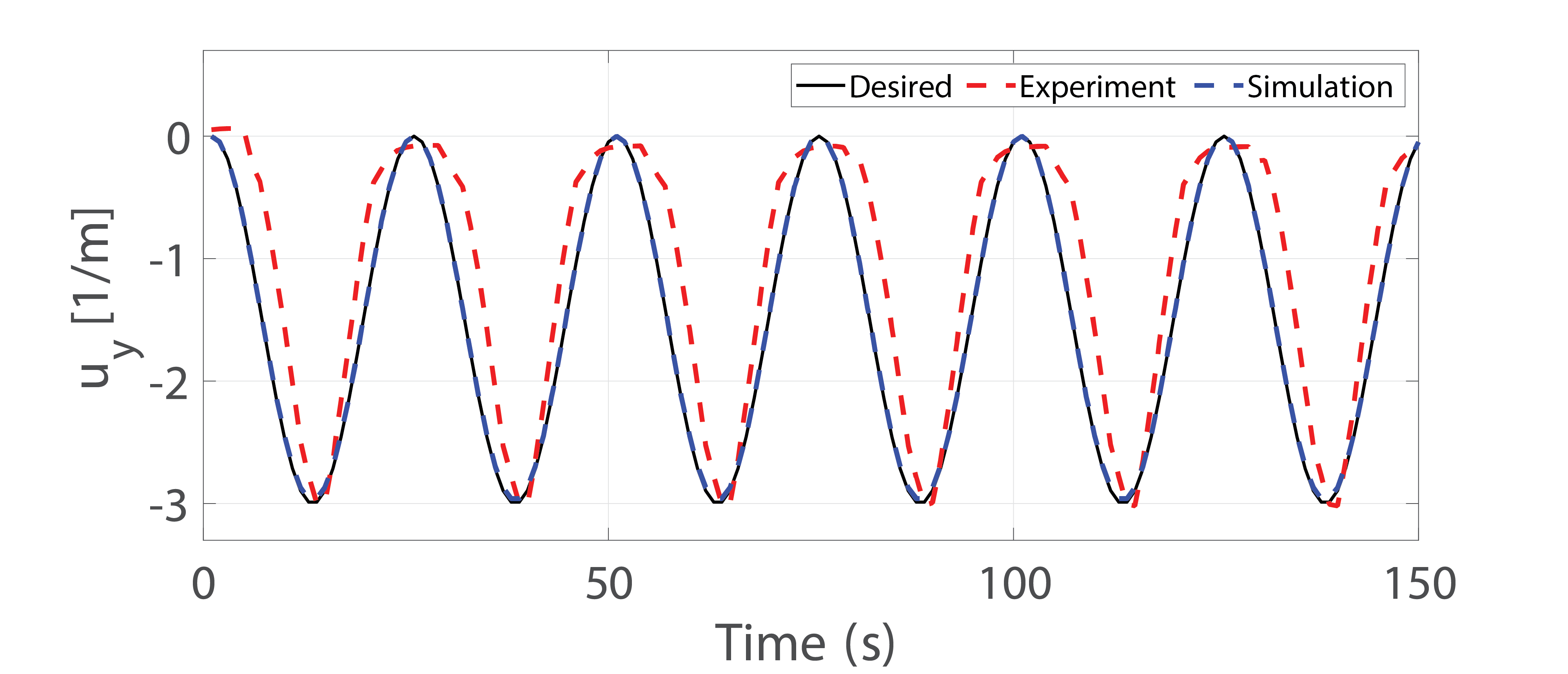} }}
    \hfill
    \subfloat[\centering $y$ coordinate of segment 1's tip in simulation and experiment. 
    ]{{\includegraphics[width=0.5\linewidth]{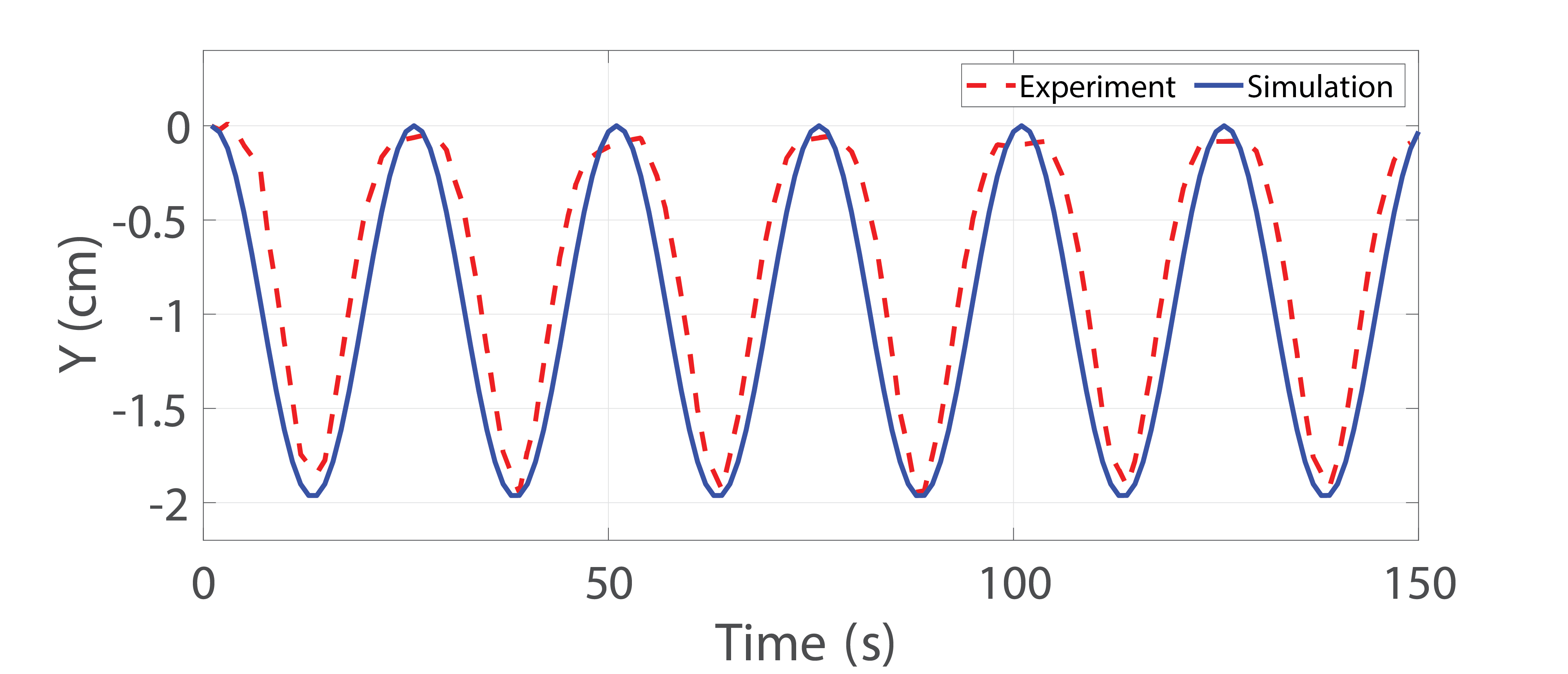} }}
    \subfloat[\centering $x$ coordinate of segment 2's tip in simulation and experiment.
    ]{{\includegraphics[width=0.5\linewidth]{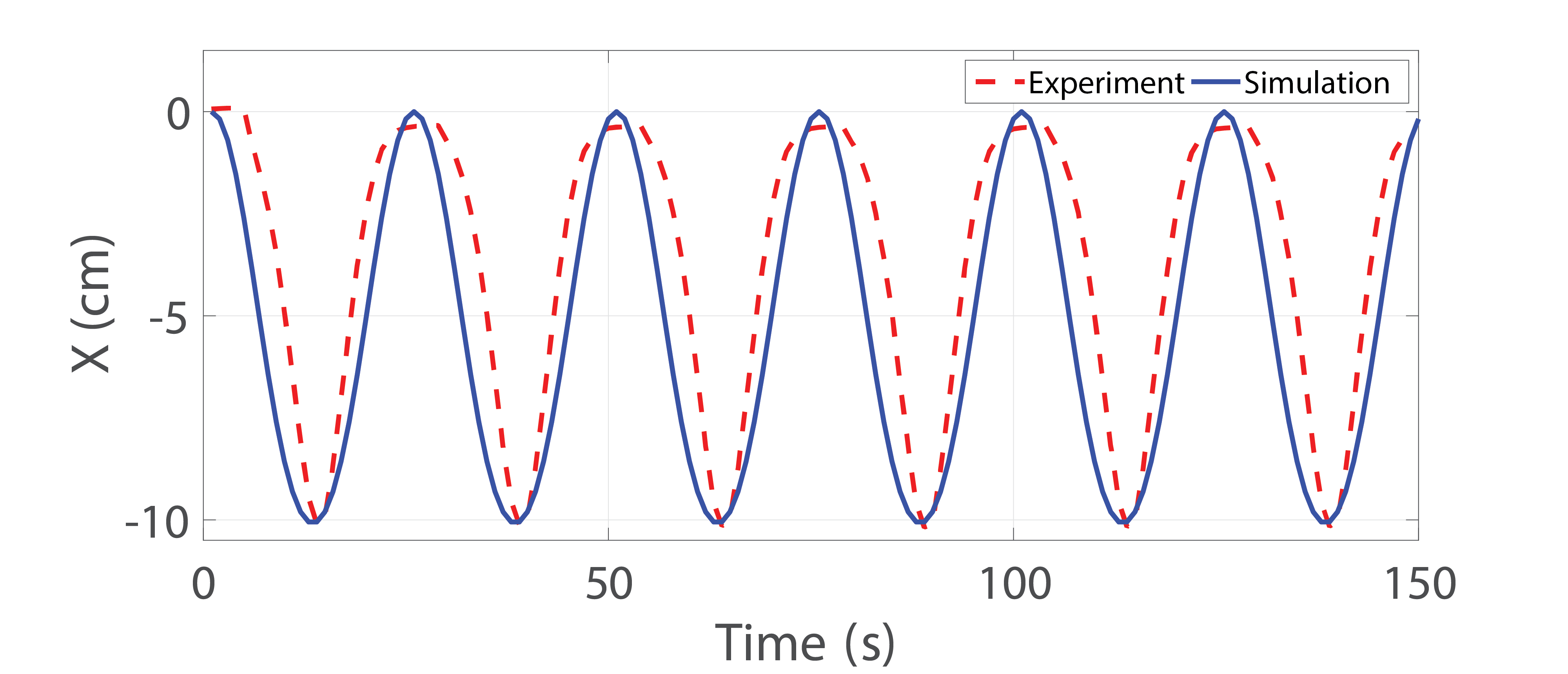} }}
    \hfill
    \subfloat[\centering Desired pressure $p_{d,j}^1$ and measured pressure $p_{m,j}^1$ for \newline 
    actuator $j$ of segment 1 during the experiment. 
    ]{{\includegraphics[width=0.5\linewidth]{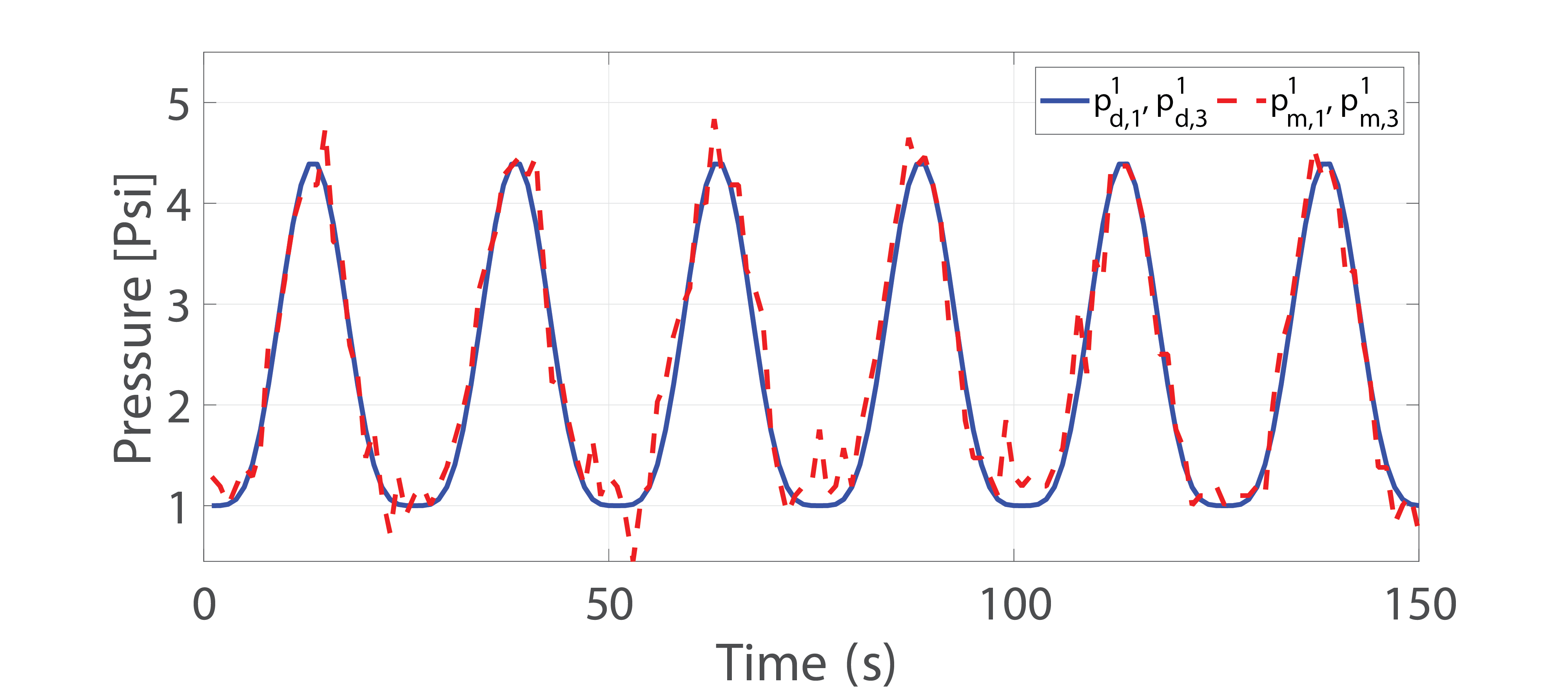} }}
    \subfloat[\centering  Desired pressure $p_{d,j}^2$ and measured pressure $p_{m,j}^2$ for \newline 
    actuator $j$ of segment 2 during the experiment. 
    ]{{\includegraphics[width=0.5\linewidth]{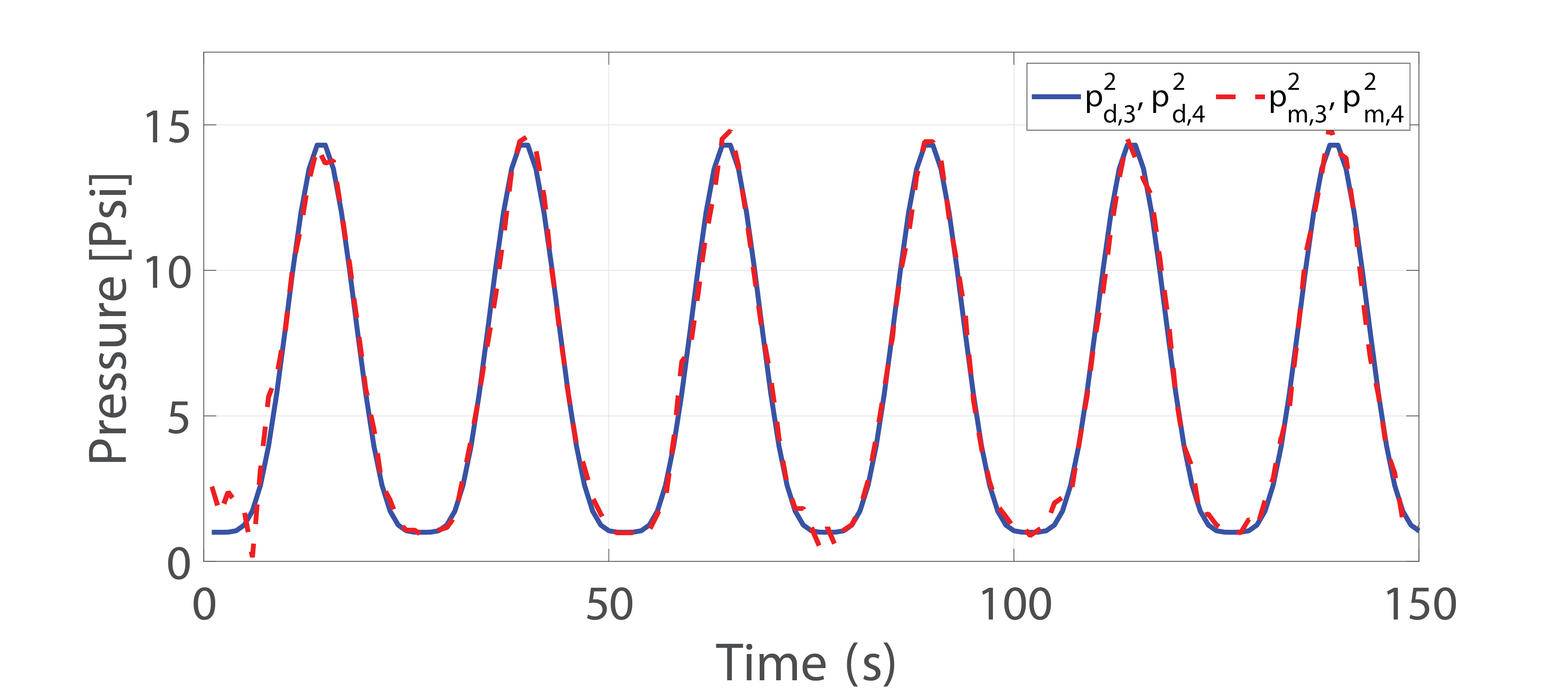} }}
    \caption{Controller performance for tracking the reference curvatures $\bar{u}_x(t)$ (segment 1) and $\bar{u}_y(t)$ (segment 2) in \eqref{Eq_u_silicone} in an experimental trial with the 
    soft robotic arm and the corresponding simulation of its dynamical model.
    }
    \label{fig:bending-control}
    \vspace{-0.2in} 
\end{figure*}
Of the reference inputs that produced bending, here we present 
the results for the input that caused segment 1 to bend about the $(+x)$-axis and segment 2 to bend about the $(-y)$-axis. This input was defined as the following curvatures:
\begin{equation}
\label{Eq_u_silicone} 
\begin{split}
&\bar{u}_{x}(t)= b \sin^2\left(\omega t\right), \quad \text{(segment 1)} \\
&\bar{u}_{y}(t)= c \sin^2\left(\omega t\right), \quad \text{(segment 2)} 
\end{split}
\end{equation}
where $b=1.5$ cm, $c=-3$ cm, $\omega=2\pi/50$ rad/s. Figure~\ref{fig:bending-control} compares the simulated and actual curvatures $u_x(t)$ and $u_y(t)$, 
and $x$ or $y$ coordinate of the segment tips, as well as the desired and measured actuator pressures. 
The robot tracks the reference input fairly well (Figs. \ref{fig:bending-control}a,b) and generally follows the prediction of the simulation (Figs.~\ref{fig:bending-control}a-d). The main source of the robot's tracking error and its discrepancy from the simulation is the delay between the pneumatic actuation and the resulting elongation or shortening of the FRAs, which was not included in the Cosserat model forward dynamics. This delay could be reduced in practice with modifications to the actuation mechanism, such as adding a vacuum to speed up deflation of the FRAs.

Table~\ref{table_RMSE_silicone} lists the root-mean-square errors (RMSEs) 
between the reference inputs and the corresponding extension or curvature of each segment during the simulations and experiments. The RMSEs were calculated from the time series plotted over 150 s in Figs.~\ref{fig:extension-control-1}a,~\ref{fig:bending-control}a, and~\ref{fig:bending-control}b. 
The highest RMSEs are the errors between the experimental and reference curvatures, which are largely due to the previously described delay between the actuation and the resulting robot deformation.

\begin{table}
\centering
\caption{RMSEs with respect to 
reference deformations.
}
\begin{tabular}{c c c c}
\hline
Deformation & Segment & Simulation & Experiment\\
\hline
$v_z$ &1 & 0.000 &  0.008\\
$v_z$ &2 &  0.000 & 0.009\\
$u_x$ &1 & 0.061 & 0.318\\
$u_y$ &2 & 0.014 & 0.662\\
\hline
\label{table_RMSE_silicone}
\end{tabular}
\end{table}
\section{Conclusions and Future Work}
\label{sec:conclusion}

In this study, a computationally efficient 
control approach that utilizes distributed sensing and actuation is implemented on a 
multi-segment soft robotic arm to track desired configurations in 3D space. The 
robotic arm can perform bending in two directions (about the $x$- and $y$-axes) and elongation in the third direction normal to the other two directions (the $z$-axis). In this control approach, the position and orientation of the tip of each segment are measured using a motion capture system and are fed back to a simulated Cosserat rod model of the robot. This model is used to estimate the unmeasured robot variables, which precludes the need for force and torque sensors on the robot, and to compute the control outputs. 
The 
local sensing and actuation capabilities of the segments allow them to be controlled independently and facilitate tracking of diverse configurations that can potentially be used for complex manipulation tasks. One feasible next step toward expanding the autonomous operations of the robot is to increase the number of its segments and incorporate embedded force sensors to enable estimation and control of the robot's interactions with its environment, similar to the proprioceptive grasping method presented in~\cite{homberg2019robust} for a soft robotic hand. 

\appendix
To assess the stability of the closed-loop system, we first substitute the expressions \eqref{Eq_control_law-silicone} for the high-level controller outputs ${}^{G}\vect{f}_p, {}^{G}\vect{l}_p$ into \eqref{Eq_force-moment-silicone}, which we then substitute into \eqref{Eq_PDE_ID} to obtain the equations:
\begin{equation}
\label{Eq_closed_3}
\begin{split}  
&{}^{G}\vect{R}\big[\vect{K}_{m_1}\vect{\bar{v}}_{tt}+\vect{K}_{v_1}(\vect{\bar{v}}_t-\vect{v}_t)+\vect{K}_{p_1}(\vect{\bar{v}}-\vect{v})\big] \\ & \quad ={}^{G}\vect{R}\rho A(\widehat{\vect{w}}\vect{q}+\vect{q}_t)-{}^{G}\vect{n}_s,\\
&{}^{G}\vect{R}\big[\vect{K}_{m_2}\vect{\bar{u}}_{tt}+\vect{K}_{v_2}(\vect{\bar{u}}_t-\vect{u}_t)+\vect{K}_{p_2}(\vect{\bar{u}}-\vect{u})\big] \\ & \quad ={}^{G}\vect{R}\rho(\widehat{\vect{w}}\vect{J}\vect{w}+\vect{J}\vect{w}_t)-{}^{G}\widehat{\vect{p}}_s{}^{G}\vect{n}-{}^{G}\vect{m}_s.
\end{split}
\end{equation}
The right-hand sides of these equations are the sums of the internal forces and moments with respect to the arc length. By defining $\vect{n'}_s$ and $\vect{m'}_s$ as the following expressions,
\begin{equation}
\label{Eq_closed_4}
\begin{split}
{}^{G}\vect{n'}_s&= {}^{G}\vect{R}\rho A(\widehat{\vect{w}}\vect{q}+\vect{q}_t)-{}^{G}\vect{n}_s, \\
{}^{G}\vect{m'}_s&= {}^{G}\vect{R}\rho(\widehat{\vect{w}}\vect{J}\vect{w}+\vect{J}\vect{w}_t)-{}^{G}\widehat{\vect{p}}_s{}^{G}\vect{n}-{}^{G}\vect{m}_s,
\end{split}
\end{equation}
and rewriting them in terms of the second time derivatives of the configuration space variables,
\begin{equation}
\label{Eq_closed_5}
\begin{split}
{}^{G}\vect{n'}_s&= {}^{G}\vect{R}\vect{K}_{m_1}\vect{v}_{tt}, \\
{}^{G}\vect{m'}_s&= {}^{G}\vect{R}\vect{K}_{m_2}\vect{u}_{tt},
\end{split}
\end{equation}
the closed-loop configuration dynamics of the robot can be expressed in the following form:
\begin{equation}
\label{Eq_closed_6}
\begin{split}  
&\vect{K}_{m_1}\vect{\bar{v}}_{tt}+\vect{K}_{v_1}(\vect{\bar{v}}_t-\vect{v}_t)+\vect{K}_{p_1}(\vect{\bar{v}}-\vect{v})=\vect{K}_{m_1}\vect{v}_{tt},\\
&\vect{K}_{m_2}\vect{\bar{u}}_{tt}+\vect{K}_{v_2}(\vect{\bar{u}}_t-\vect{u}_t)+\vect{K}_{p_2}(\vect{\bar{u}}-\vect{u})=\vect{K}_{m_2}\vect{u}_{tt}.
\end{split}
\end{equation}
Defining the error vector $\vect{e}(t)=(\vect{\bar{v}}-\vect{v},\vect{\bar{u}}-\vect{u})^T$
and writing \eqref{Eq_closed_6} in terms of this error, the closed-loop system dynamics take the form of standard homogeneous second-order differential equations:
\begin{equation}
\label{Eq_error}
\begin{split}
\vect{e}_{{tt}}+\vect{K'}_{v}\vect{e}_{{t}}+\vect{K'}_{p}\vect{e}=\vect{0},
\end{split}
\end{equation}
where the matrices $\vect{K'}_{v}$ and $\vect{K'}_{p}$ are defined as
\begin{equation}
\label{Eq_matrices_MKK}
\begin{split}
\vect{K'}_v&=\begin{bmatrix} \vect{K}_{v_1} \oslash \vect{K}_{m_1} & 0 \\ 0 & \vect{K}_{v_2} \oslash \vect{K}_{m_2}\end{bmatrix},\\
\vect{K'}_p&=\begin{bmatrix} \vect{K}_{p_1} \oslash \vect{K}_{m_1} & 0 \\ 0 & \vect{K}_{p_2} \oslash \vect{K}_{m_2}\end{bmatrix},
\end{split}
\end{equation}
\noindent in which $\oslash$ denotes element-wise division of matrices (Hadamard division).
The matrices $\vect{K'}_{v}$ and $\vect{K'}_{p}$  are symmetric and positive definite. 
Next, the following positive definite quadratic Lyapunov function is chosen,
\begin{equation}
\label{Eq_Lyapunov}
\begin{split}
V=\dfrac{1}{2}\vect{e}_{t}^T\vect{e}_{t}+\dfrac{1}{2}\vect{{e}}^T\vect{K'}_{p}\vect{e},
\end{split}
\end{equation}
\noindent which has the following time derivative:
\begin{equation}
\label{Eq_Lyapunov_d}
\begin{split}
V_{t}&=\dfrac{1}{2}\vect{e}_{tt}^T\vect{e}_{t}+\dfrac{1}{2}\vect{e}_{t}^T\vect{e}_{tt}+\dfrac{1}{2}\vect{e}_{t}^T\vect{K'}_{p}\vect{e}+\dfrac{1}{2}\vect{e}^T\vect{K'}_{p}\vect{e}_{t}\\
&=\dfrac{1}{2}(\vect{e}_{tt}^T+\vect{e}^T\vect{K'}_{p})\vect{e}_{t}+\dfrac{1}{2}\vect{e}_{t}^T(\vect{e}_{tt}+\vect{K'}_{p}\vect{e})\\
&=\dfrac{1}{2}(-\vect{e}_{t}^T\vect{K'}_{v})\vect{e}_{t}+\dfrac{1}{2}\vect{e}_{t}^T(-\vect{K'}_{v}\vect{e}_{t})=-\vect{e}_t^T\vect{K'}_{v}\vect{e}_{t}.
\end{split}
\end{equation}
Since $\vect{K'}_{v}$ is positive definite, $V_t$ is a negative definite function. 
By applying Lyapunov's direct method to the closed-loop system dynamics, we can prove that $\vect{e}(t)\xrightarrow{}\vect{0}$ as $t\xrightarrow{}\infty$ and the system is globally asymptotically stable~\cite{khalil2002nonlinear}.

\bibliographystyle{IEEEtran}
\bibliography{RA-L-2022.bib}

\end{document}